\documentclass[10pt,twocolumn,letterpaper]{article}
\newif\ifarxiv\arxivtrue
\ifarxiv\usepackage[pagenumbers]{cvpr}\else\usepackage[review]{cvpr}\fi
\ifdefined\XeTeXrevision
  \usepackage{fontspec}
\fi
\usepackage{graphicx,booktabs,multirow,amsmath,amssymb,microtype,xspace,enumitem}
\definecolor{cvprblue}{rgb}{0.21,0.49,0.74}
\usepackage[breaklinks,colorlinks,allcolors=cvprblue]{hyperref}
\newcommand{\method}{\textsc{X-Fly}\xspace}
\newcommand{\up}{$\uparrow$}
\newcommand{\dn}{$\downarrow$}
\def\paperID{*****}\def\confName{CVPR}\def\confYear{2027}
\title{\method: Learning to Reach and Stop in\\Full-Chunk Aerial Vision-Language Navigation}
\ifarxiv
\author{%
  Ruibin Ju\textsuperscript{*}\qquad Ruifeng Cai\qquad Min Xu\qquad Zhuojie Chen\qquad Lin Wang\\[1pt]
  {\normalsize\raisebox{-0.12em}{\includegraphics[height=0.95em]{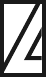}}\hspace{0.35em}ARIDGE, XPENG}\\[-1pt]
  {\small\textsuperscript{*}Corresponding author: {\hypersetup{urlcolor=black}\href{mailto:jurb@aridge.com}{jurb@aridge.com}}}\\[-1pt]
  {\small Project page: \url{https://juliusju572.github.io/x-fly/}}%
}
\hypersetup{pdftitle={X-Fly: Learning to Reach and Stop in Full-Chunk Aerial Vision-Language Navigation},pdfauthor={Ruibin Ju, Ruifeng Cai, Min Xu, Zhuojie Chen, Lin Wang}}
\else
\author{Anonymous CVPR Submission}
\fi
\begin{document}
\maketitle

\begin{abstract}
Aerial vision-language navigation (VLN) agents built on large vision-language models are expensive to query. Executing a complete predicted action chunk before replanning amortizes this cost, reducing model inference from 195 to 42\,ms per executed waypoint for our policy. It also moves the states at which the policy is queried away from expert demonstrations, which supervise neither recovery from these states nor termination at the learner's own arrival. On OpenFly test-seen, no-stop failures account for 47.78\% of full-chunk behavior cloning (BC) outcomes. We present \method, a staged training framework for full-chunk aerial VLN. Recovery DAgger labels learner-visited decision states with smooth chunks that reconnect to future expert segments and with STOP at the first in-region state. Flow-GRPO then jointly refines the chunked flow controller and a Bernoulli STOP policy from same-task closed-loop outcomes, supplemented by outcome-verified stopping supervision. With eight controls per decision, \method reaches 74.18\% SR on test-seen and 51.73\% on test-unseen, compared with 16.45\% and 18.43\% for full-chunk BC and 32.04\% and 29.46\% for DreamFly.
\end{abstract}

\section{Introduction}
\begin{figure}[b]
  \centering
  \includegraphics[width=\columnwidth]{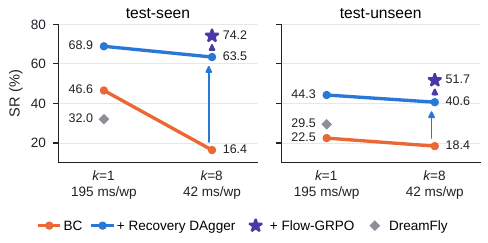}
  \caption{\textbf{Accuracy--inference trade-off of \method stages.} SR under single-step ($k{=}1$) and full-chunk ($k{=}8$) execution; x-labels give amortized model inference per executed waypoint. On test-seen, full-chunk BC reaches the goal but often does not stop. Recovery DAgger makes full-chunk execution viable, and Flow-GRPO, trained and evaluated at $k{=}8$ only, gives the highest SR among the evaluated operating points at 4.7$\times$ less amortized model inference time per executed waypoint than single-step execution. Inference timings apply to the \method backbone; DreamFly~\cite{deng2026dreamfly} is shown as an SR reference at $k{=}1$.}
  \label{fig:teaser}
\end{figure}
Aerial VLN asks an embodied agent to follow a free-form instruction through a large urban environment and terminate within a target radius~\cite{liu2023aerialvln,gao2025openfly}. Compared with indoor graph navigation~\cite{anderson2018vln,chen2021hamt}, the agent must produce continuous translation and rotation, cope with large viewpoint changes, and recover from control errors that alter every subsequent observation. Success therefore depends on two coupled abilities: flying to the described goal, and deciding when the goal has been reached.

Recent aerial agents condition on large vision-language models (VLMs), which makes every policy query expensive~\cite{deng2026dreamfly,zhu2026fsdvln}. Action-chunk policies predict several future controls per query~\cite{zhao2023act,chi2024diffusionpolicy}, yet aerial VLN agents such as DreamFly execute only the first control and query again~\cite{deng2026dreamfly}. Executing the complete chunk amortizes policy inference across its controls: for our fast--slow policy, executing all eight cuts model inference from 195 to 42\,ms per executed waypoint (Sec.~\ref{sec:setup}).

Full-chunk execution also changes where the policy is queried. The aircraft moves through eight controls before new visual feedback can correct it, so the states at which the policy is next queried, its \emph{decision anchors}, drift away from the expert trajectories it was trained on. Expert-only data cover these learner-visited anchors poorly in two ways: they contain no chunk that starts at an off-route learner pose and returns it to the route, and their positive STOP labels are confined to expert terminal windows, leaving the anchors where the learner itself arrives poorly represented. We call this the \emph{full-chunk support gap}. On test-seen, switching BC from single-step to full-chunk execution drops SR from 46.57\% to 16.45\% (Fig.~\ref{fig:teaser}), although navigation failures barely change (28.02\% vs.\ 30.77\%); no-stop failures account for 47.78\% of the full-chunk outcomes (Sec.~\ref{sec:failure}). Reaching the goal is not the same as stopping there.

\begin{figure*}[t]
  \centering
  \includegraphics[width=\textwidth]{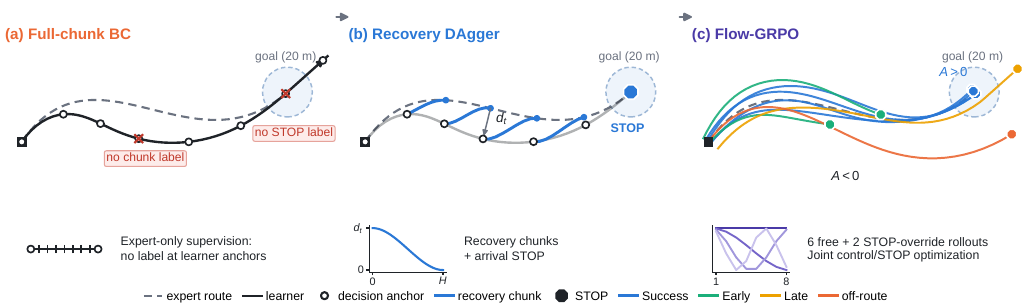}
  \caption{\textbf{The full-chunk support gap and how \method addresses it.} (a)~Full-chunk BC flies eight controls between decisions and visits learner-induced anchors that expert demonstrations cover poorly, with neither a corrective chunk nor an arrival STOP label; the policy overshoots the goal. (b)~Recovery DAgger relabels each learner anchor with a chunk whose offset $d_t$ decays by a smoothstep (inset) and rejoins the expert route at the chunk end; the first in-region anchor receives STOP. (c)~Flow-GRPO samples six free and two STOP-override attempts per task with low-frequency chunk noise (inset: the four cosine modes) and a Bernoulli STOP policy, and uses their relative outcomes to update both.}
  \label{fig:training}
\end{figure*}

\method addresses both sides of the gap with two training stages applied after multi-domain BC (Fig.~\ref{fig:training}). \emph{Recovery DAgger}~\cite{ross2011dagger} runs the learner with the deployed full-chunk executor and labels every visited anchor with a smooth chunk that decays the learner's offset and reconnects to the future expert segment; the first anchor inside the success region receives a positive STOP label. Recovery reduces navigation and no-stop failures, while premature stopping remains common. \emph{Flow-GRPO}~\cite{liu2025flowgrpo} then compares closed-loop attempts of the same task and updates the chunked flow controller and a separate Bernoulli STOP policy from their outcomes. With eight controls per decision, \method reaches 74.18\% SR on test-seen and 51.73\% on test-unseen, all from explicit in-region STOP.

Our contributions are:
\begin{itemize}[leftmargin=1.25em,itemsep=1pt,topsep=2pt]
  \item We study full-chunk execution in aerial VLN and characterize the support gap it creates at learner-visited decision anchors, with separate effects on goal reaching and termination.
  \item We present \method, a staged training framework for full-chunk aerial navigation that combines multi-domain BC, recovery supervision at learner-visited anchors, and closed-loop outcome optimization of chunked control and STOP.
  \item We demonstrate improved full-chunk navigation on OpenFly and characterize distinct stage-level gains: recovery improves goal-region reachability and reduces no-stop failures, while outcome post-training further reduces premature termination.
\end{itemize}

\section{Related Work}
\paragraph{Aerial vision-language navigation.}
VLN was first studied on discrete indoor navigation graphs~\cite{anderson2018vln,fried2018speaker,wang2019rcm,hong2021vlnbert,chen2021hamt,chen2022duet}. Continuous VLN removes the graph abstraction~\cite{krantz2020beyond}, and AerialVLN and OpenFly extend the setting to long-range urban flight~\cite{liu2023aerialvln,gao2025openfly}. Recent aerial agents use hierarchical planning, causal memory, diffusion policies, world-action prediction, and fast--slow reasoning~\cite{wang2025traveluav,zhang2025citynavagent,deng2026dreamfly,zhao2026worldvln,liu2026imagineuav,zhu2026fsdvln}. We adopt a fast--slow backbone and study a different axis: the training distribution shift created when such a policy executes a complete chunk before replanning.

\paragraph{Learning when to stop.}
Graph VLN agents treat STOP as one action of a discrete vocabulary, and several works show that termination deserves dedicated modeling, through a separate stop policy~\cite{xiang2020stop} or progress estimation~\cite{ma2019selfmonitoring}. DreamFly adds a lightweight termination head trained on top of its frozen navigation policy~\cite{deng2026dreamfly}. A continuous-waypoint policy has no intrinsic STOP coordinate, so \method also uses a separate Bernoulli head, but supervises it at the learner's own arrival anchors and optimizes it jointly with control from closed-loop outcomes.

\paragraph{Flow policies and action chunks.}
Diffusion and flow-matching policies represent multimodal continuous actions and emit action sequences~\cite{chi2024diffusionpolicy,lipman2023flow,black2025pi0}. Action chunking amortizes policy inference~\cite{zhao2023act}; temporal ensembling, bidirectional decoding, and real-time chunking address the resulting loss of reactivity at inference time~\cite{zhao2023act,liu2025bid,black2025rtc}. Fast--slow VLA systems separate reasoning from control~\cite{xu2025vlacache,li2026think}. \method keeps inference unchanged and instead repairs the training distribution that full-chunk execution induces.

\paragraph{Closed-loop imitation and reinforcement learning.}
DAgger reduces covariate shift by labeling learner-visited states~\cite{ross2011dagger}. DART injects noise into demonstrations~\cite{laskey2017dart}, HG-DAgger relies on a human corrector~\cite{kelly2019hgdagger}, and ChauffeurNet synthesizes perturbed driving trajectories that return to the route~\cite{bansal2019chauffeurnet}. TravelUAV applies continuous-trajectory DAgger to aerial navigation with a teacher that searches forward along the reference~\cite{wang2025traveluav}. Our labels are instead collected at the anchors produced by full-chunk execution, follow the route by the number of executed controls, reconnect within one chunk, and add STOP labels at the learner's arrival anchors. On the RL side, RCM and EnvDrop combine imitation with RL for VLN~\cite{wang2019rcm,tan2019envdrop}, RefineFly and WorldVLN post-train aerial policies in closed loop~\cite{wang2026recoverfly,zhao2026worldvln}, and diffusion and flow policies are optimized by treating denoising as a Markov decision process~\cite{black2024ddpo,ren2025dppo,liu2025flowgrpo,zhang2025reinflow}. We use Flow-GRPO's stochastic transitions and add a factorized STOP objective to refine recovered full-chunk behavior.

\section{Method}
\subsection{Problem: the full-chunk support gap}
\label{sec:problem}
An episode consists of an instruction, onboard RGB observations, and executed controls, and ends at a STOP decision, the control budget, or an abnormal event such as a collision. The fused policy is queried only at \emph{decision anchors} $s_t$, one every $k$ executed controls. At each anchor it outputs a control chunk $U_t=(u_{t,1},\ldots,u_{t,H})$ with $H=8$ and a STOP decision $z_t\in\{0,1\}$. If $z_t=0$, the first $k$ controls of $U_t$ are executed open loop before the next anchor. We deploy $k=H$.

BC supervises $(U,z)$ only at anchors sampled from expert trajectories, $s\sim d^{\star}$, whereas deployment queries the policy at $s\sim d^{\pi}_{k}$, the anchor distribution induced by the learner's own $k$-step execution. Larger $k$ tends to move $d^{\pi}_{k}$ farther from $d^{\star}$, because more controls separate consecutive corrections. Two supervision signals are then missing on the support of $d^{\pi}_{k}$:
\begin{itemize}[leftmargin=1.25em,itemsep=1pt,topsep=2pt]
  \item \textbf{Control gap.} At an off-route anchor, no expert chunk starts at the learner's pose and remains coherent for $H$ steps until the next decision.
  \item \textbf{Termination gap.} Positive STOP labels come only from expert terminal windows, not from the anchor where the learner itself first enters the success region.
\end{itemize}
BC (Sec.~\ref{sec:bc}) provides the initial policy, Recovery DAgger (Sec.~\ref{sec:recovery}) supplies both missing labels at the anchors this policy visits at $k{=}H$, and Flow-GRPO (Sec.~\ref{sec:grpo}) refines the recovered policy from closed-loop outcomes.

\subsection{Policy and BC initialization}
\label{sec:bc}
\begin{figure*}[t]
  \centering
  \includegraphics[width=\textwidth]{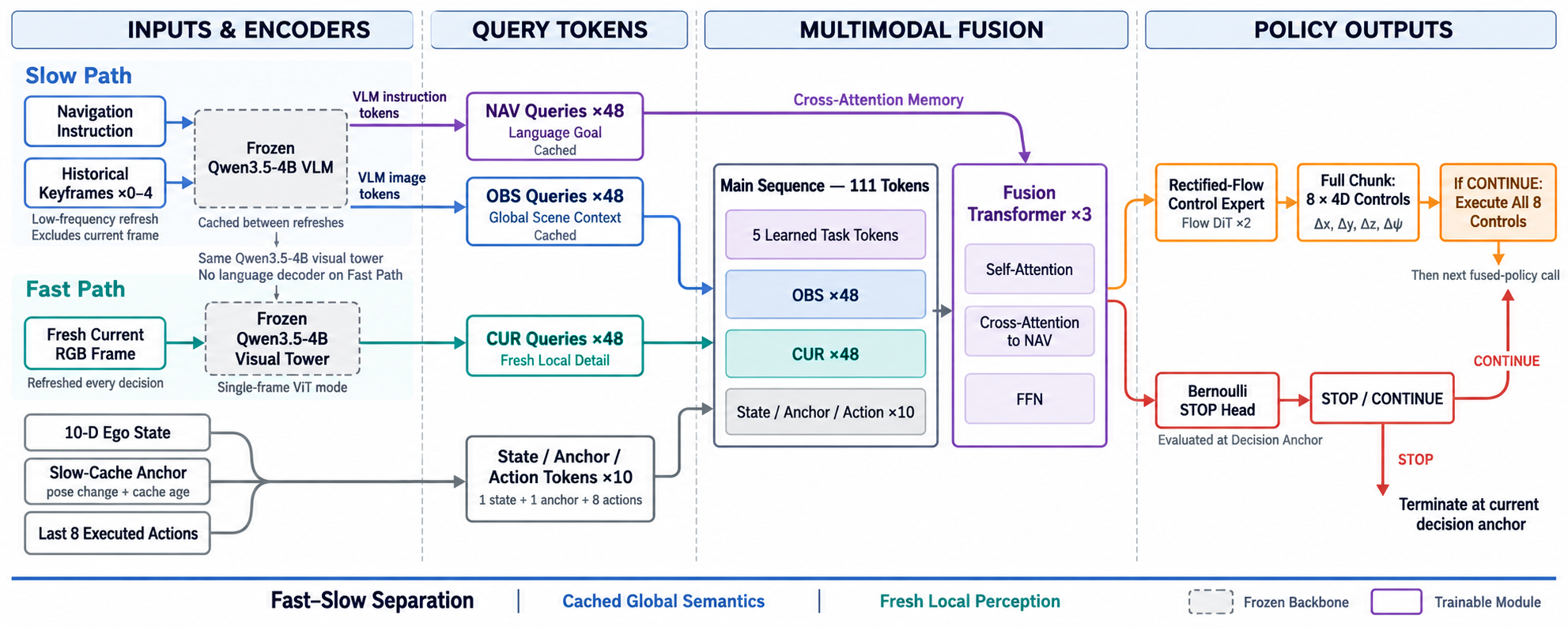}
  \caption{\textbf{Policy backbone shared by the three training stages.} A slow path runs the frozen Qwen3.5-4B VLM on the instruction and causal keyframes; a fast path runs only its visual tower on the current frame. Fused with ten state/anchor/action tokens (ego state, slow-cache anchor, and the last eight executed actions), the representation drives a rectified-flow controller that emits eight 4-D controls $(\Delta x,\Delta y,\Delta z,\Delta\psi)$ and a separate Bernoulli STOP head; if CONTINUE is sampled, all eight controls are executed.}
  \label{fig:policy}
\end{figure*}

The policy (Fig.~\ref{fig:policy}) follows the fast--slow design of recent aerial agents~\cite{deng2026dreamfly,zhu2026fsdvln}. Learned query banks extract cached observation and instruction features from a frozen VLM and fresh current-frame features from its visual tower; these tokens, ego state, and a masked history of executed controls form the fusion sequence, with navigation queries as cross-attention memory. The fused representation conditions a rectified-flow controller and a separate STOP head, so control and termination have distinct likelihoods.

Each control $u_{t,j}=(\Delta x,\Delta y,\Delta z,\Delta\psi)$ is a body-frame increment of translation and yaw. A two-layer DiT is trained by rectified flow~\cite{lipman2023flow}: with expert chunk $U^\star$, noise $\epsilon$, and $X_\tau=(1-\tau)\epsilon+\tau U^\star$,
\begin{equation}
  \mathcal L_{\mathrm{flow}}=\mathbb E\!\left[\|v_\theta(X_\tau,\tau,c)-(U^\star-\epsilon)\|_2^2\right],
\end{equation}
where $c$ is the fused conditioning. Six Euler steps generate a chunk at inference. BC trains the query banks, fusion layers, controller, and STOP head on 849,106 valid decision windows from the UE, 3DGS, and AirSim domains of OpenFly, with positive STOP labels at expert terminal windows; the VLM and visual tower stay frozen. Auxiliary regularizers are listed in Appendix~\ref{sec:impl}.

\subsection{Recovery DAgger at learner-visited anchors}
\label{sec:recovery}
We run the current learner in training scenes with the deployed full-chunk executor and record observations at its anchors. Labels must follow demonstration order, since a nearest-point match can point backward when a route passes near itself. We therefore index the expert route with a monotonic cursor $c_t$ that advances at each collected decision by the number of controls actually executed, rather than searching forward along the reference for a match as in TravelUAV~\cite{wang2025traveluav}. With learner pose $p_t$ at anchor $t$ and offset $d_t=p_t-p^{\star}_{c_t}$, the recovery target is
\begin{equation}
  p_{t,j}^{\mathrm{rec}}=p_{c_t+j}^{\star}+d_t\,(1-3q_j^2+2q_j^3),\quad q_j=j/H,
  \label{eq:recovery}
\end{equation}
for $j=0,\ldots,H$. The offset factor equals one at $j{=}0$, decays with zero slope at both ends, and vanishes at $j{=}H$, so the target starts at the learner pose and rejoins the future expert segment after one chunk (Fig.~\ref{fig:training}b). Yaw uses the same decaying offset, wrapped to the shortest signed angle, and consecutive target poses are converted to body-frame controls as in BC. Recovery targets are constructed only during training.

STOP labels follow goal-region entry rather than the cursor. Collection runs with STOP disabled and is truncated at the first anchor inside the 20\,m success region, which is labeled STOP with a hover target; every earlier anchor is CONTINUE with its recovery target, and an anchor still outside the region after the cursor has reached the expert endpoint keeps a target toward the goal. Truncation also removes post-arrival anchors whose labels would point backward. Recovery thus adds STOP positives at the learner's own arrival anchors, complementing the expert terminal windows used by BC. Collection and training alternate for three rounds, the first initialized from BC and later rounds from the preceding recovery checkpoint, each mixing recovery windows 1:1 with BC windows (Appendix~\ref{sec:impl}).

\subsection{Outcome optimization with Flow-GRPO}
\label{sec:grpo}
Recovery improves goal-region reachability and reduces no-stop failures, while premature termination remains common (Sec.~\ref{sec:failure}). The final stage starts from the recovered full-chunk policy and uses actual task outcomes to refine termination timing together with chunked control. After recovery, closed-loop attempts of the same task differ in route progress, goal entry, overshoot, and termination, so same-task outcomes provide relative advantages without a learned value function. We build on Flow-GRPO~\cite{liu2025flowgrpo} and instantiate it for full-chunk aerial control.

\paragraph{Rollout groups.}
For one instruction and start pose, a frozen copy of the policy produces a group of eight closed-loop attempts that share the initial flow latent at each decision depth but sample independent SDE-tail and STOP noise. Six are free attempts of the deployed policy. Free attempts often stop at similar depths, which gives the STOP term little contrast, so the remaining two are training-only STOP-override attempts that override their first $n\in\{1,2,3\}$ sampled STOP events to CONTINUE and then run freely. Let $\mathcal A_t$ be the attempts with a decision at depth $t$ and $R^{\mathrm{go}}_{i,t}$ their return-to-go. The advantage is
\begin{equation}
  A_{i,t}=\frac{R^{\mathrm{go}}_{i,t}-\mu_t}{\max(s_t,s_{\mathrm{floor}})},\quad i\in\mathcal A_t,
  \label{eq:advantage}
\end{equation}
with $\mu_t,s_t$ the mean and standard deviation over $\mathcal A_t$ and $s_{\mathrm{floor}}=0.5$. Normalizing within a task group and depth controls for route difficulty and reward scale. Override outcomes enter these statistics, but their decisions do not enter the policy-gradient term below.

\paragraph{Objective.}
Let $\pi_\theta^s$ be the density of stochastic denoising step $s$ and $\pi_\theta^z$ the Bernoulli probability of the sampled STOP/CONTINUE action under the training-time STOP temperature 1.5. Each rollout batch is used for one KL-regularized policy-gradient update over free-attempt decisions,
\begin{equation}
\begin{split}
\mathcal L_{\mathrm{actor}}={}&-\mathbb E\Bigl[A_t\Bigl(\textstyle\sum_{s=3}^{5}\log\pi_\theta^s+\log\pi_\theta^z\Bigr)\Bigr]\\
&+\beta(D_{\mathrm{KL}}^u+D_{\mathrm{KL}}^z),
\end{split}
\label{eq:actor}
\end{equation}
where the control term is applied to decisions that execute a chunk, the STOP term to legal decisions (STOP is legal from the second decision on), and both KL terms regularize toward the frozen initial policy with $\beta=0.01$. At the rollout policy, its gradient equals that of the clipped Flow-GRPO surrogate. The override attempts additionally supply outcome-verified stopping labels: a legal decision inside the region is a positive, and an overridden STOP outside the region is a negative only if the continued attempt later reaches the region. A margin-ranking and BCE loss $\mathcal L_{\mathrm{rank}}$ on same-depth pairs completes the stage objective $\mathcal L_{\mathrm{actor}}+0.5\mathcal L_{\mathrm{rank}}$ (Appendix~\ref{sec:impl}).

\paragraph{Reward.}
The reward is dominated by the task outcome:
\begin{equation}
\begin{split}
R={}&R_{\mathrm{task}}+2\,\mathbb I[\mathrm{first\ entry}]-1.5\,\mathbb I[\mathrm{leave\ after\ entry}]\\
&+R_{\mathrm{progress}}+R_{\mathrm{control}}-\kappa\,\mathbb I[\mathrm{unsuccessful}].
\end{split}
\label{eq:reward}
\end{equation}
An episode is unsuccessful iff its final position lies outside the success region. A STOP inside the region receives $R_{\mathrm{task}}=+20$, and an unsuccessful episode that ends at a non-STOP terminal condition receives $-10$. $R_{\mathrm{progress}}$ gives capped credit for approaching the goal, $R_{\mathrm{control}}$ penalizes invalid, misdirected, or abrupt controls, and the constant shift $\kappa$ keeps every unsuccessful return negative (Appendix~\ref{sec:impl}).

\paragraph{Stochastic chunk transitions.}
We introduce stochastic denoising transitions during training to obtain tractable conditional likelihoods. As in Flow-GRPO, the last three of six Euler denoising steps, $s\in\{3,4,5\}$, are stochastic actions of an augmented MDP whose state is the partial chunk and its conditioning~\cite{black2024ddpo,liu2025flowgrpo}. We place the exploration noise in a low-frequency temporal basis: with $\Phi\in\mathbb R^{H\times M}$ holding $M=4$ orthonormal cosine modes over the $H=8$ waypoints and $\Psi$ its orthonormal complement, each stochastic step samples
\begin{equation}
\begin{split}
x^{s+1}&=\mu_\theta^s+\nu_s(\Phi\epsilon_s+\rho\Psi\xi_s),\\
C_s&=\nu_s^2(\Phi\Phi^\top+\rho^2\Psi\Psi^\top)\otimes I_D,
\end{split}
\label{eq:sde}
\end{equation}
with $\epsilon_s,\xi_s$ standard Gaussian, $D=4$ control dimensions, and $\rho=0.1$. Low-frequency variation dominates exploration, so sampled chunks remain smooth, while the small complement noise keeps $C_s$ full rank. Because $C_s$ is diagonal in the $(\Phi,\Psi)$ basis, each transition has a closed-form log-density over the complete $H\times D$ chunk; the restoring mean $\mu_\theta^s$, noise schedule $\nu_s$, and density are given in Appendix~\ref{sec:density}. Deployment keeps the deterministic six-step solver.

\section{Experiments}
\subsection{Setup}
\label{sec:setup}
We evaluate on the DreamFly-processed OpenFly splits~\cite{gao2025openfly,deng2026dreamfly}: test-seen has 1,392 trajectories from UE BigCity and six AirSim cities, and test-unseen has 404 trajectories from the held-out UE SmallCity. Following the benchmark, an episode succeeds if its final position lies within 20\,m of the goal. Here $k$ is the number of controls executed per fused-policy decision; full-chunk execution is $k{=}8$.

\paragraph{Evaluating a stochastic STOP policy.}
At each legal decision, \method samples $z_t\sim\operatorname{Bernoulli}(p_t)$ with $p_t=\sigma(\ell_t)$ from the STOP logit $\ell_t$, \ie at unit temperature, the same temperature at which BC and Recovery DAgger train the STOP head. Flow-GRPO samples STOP at temperature 1.5 only during training rollouts for exploration (Appendix~\ref{sec:impl}); all reported \method results use $T{=}1$. We record each route once with STOP disabled and analytically marginalize the Bernoulli STOP decisions conditional on this control rollout. The probability of first stopping at decision $t$ is $p_t\prod_{j<t}(1-p_j)$, and the residual $\prod_j(1-p_j)$ is assigned to the full rollout. SR, OSR, SPL, and navigation error (NE) are expectations under this first-stop distribution, each computed on the executed prefix up to the stop. Every \method row uses this rule; prior methods are reported with their native stopping policies. The same recordings also give two diagnostics: \emph{anchor reach} (some decision anchor lies inside the region) and \emph{dense reach} (some executed waypoint does).

\paragraph{Amortized inference cost.}
We profile the deployed policy on an NVIDIA RTX PRO 5000 72GB GPU. The slow VLM path refreshes once per 16 executed waypoints at 0.320\,s per refresh; each fresh-frame visual-tower call costs 0.141\,s and each fusion/control/STOP call 0.034\,s. Over a 16-waypoint window, $k=1$ costs $0.320+16(0.141+0.034)=3.12$\,s and $k=8$ costs $0.320+2(0.141+0.034)=0.67$\,s, \ie 195 vs.\ 42\,ms per executed waypoint (4.66$\times$).

\subsection{Comparison with prior work}
\begin{table*}[t]
\centering
\caption{OpenFly test-seen and test-unseen results under the DreamFly-processed benchmark. NE is in meters, other metrics in \%. Prior results are as reported with their native stopping policies; every \method row uses the same 128-control cap and Bernoulli STOP policy, evaluated over its first-stop distribution (Sec.~\ref{sec:setup}). $k$: controls executed per fused-policy decision. The two Recovery DAgger rows evaluate the same checkpoint, and its $k{=}8$ operating point initializes Flow-GRPO.}
\label{tab:main}
\resizebox{\textwidth}{!}{\begin{tabular}{lrrrrrrrr}
\toprule
& \multicolumn{4}{c}{test-seen} & \multicolumn{4}{c}{test-unseen}\\
\cmidrule(lr){2-5}\cmidrule(lr){6-9}
Method & SR \up & OSR \up & SPL \up & NE \dn & SR \up & OSR \up & SPL \up & NE \dn\\
\midrule
Random & 13.51 & 18.75 & 9.72 & 65.67 & 15.35 & 23.76 & 11.31 & 59.99\\
Action Sampling & 15.95 & 26.51 & 13.67 & 62.78 & 20.54 & 32.67 & 17.22 & 55.27\\
Seq2Seq & 24.35 & 61.93 & 19.35 & 54.44 & 26.49 & 61.88 & 19.62 & 47.69\\
CMA & 7.97 & 69.32 & 6.26 & 313.03 & 5.69 & $\mathbf{73.02}$ & 3.92 & 230.05\\
AerialVLN~\cite{liu2023aerialvln} & 16.52 & 65.66 & 14.63 & 176.29 & 9.65 & 68.07 & 7.93 & 161.19\\
OpenFly-Agent~\cite{gao2025openfly} & 22.63 & 52.73 & 20.42 & 122.89 & 14.11 & 62.38 & 12.49 & 163.87\\
DreamFly~\cite{deng2026dreamfly}, $k{=}1$ & 32.04 & 46.77 & 28.22 & 44.87 & 29.46 & 46.78 & 23.54 & 45.29\\
\midrule
\method BC, $k{=}1$ & 46.57 & 52.85 & 39.92 & 37.28 & 22.49 & 23.46 & 21.00 & 58.61\\
\method BC, $k{=}8$ & 16.45 & 67.30 & 13.54 & 37.87 & 18.43 & 43.14 & 16.18 & 48.12\\
\quad + Recovery DAgger, $k{=}1$ & 68.88 & 75.45 & 56.84 & 20.12 & 44.27 & 47.64 & 39.96 & 36.43\\
\quad + Recovery DAgger, $k{=}8$ & 63.46 & 72.87 & 54.54 & 21.49 & 40.63 & 52.71 & 36.46 & 35.16\\
\quad + Flow-GRPO (\textbf{\method, full}), $k{=}8$ & $\mathbf{74.18}$ & $\mathbf{84.07}$ & $\mathbf{66.34}$ & $\mathbf{17.77}$ & $\mathbf{51.73}$ & 68.07 & $\mathbf{47.03}$ & $\mathbf{29.17}$\\
\bottomrule
\end{tabular}}
\end{table*}

With eight controls per decision, \method reaches 74.18\% SR, 66.34\% SPL, and 17.77\,m NE on test-seen, and 51.73\% SR, 47.03\% SPL, and 29.17\,m NE on test-unseen, improving all four metrics over DreamFly on both splits (Table~\ref{tab:main}). Within the same backbone and execution setting, Recovery DAgger raises SR from 16.45\% to 63.46\% on test-seen and from 18.43\% to 40.63\% on test-unseen, and Flow-GRPO adds a further 10.72 and 11.10 points. On test-unseen, single-step BC on our backbone trails DreamFly (22.49\% vs.\ 29.46\%), and the two closed-loop stages provide the advantage. Flow-GRPO updates draw only BigCity training tasks, yet its gain carries over to the held-out SmallCity.

\subsection{Failure attribution}
\label{sec:failure}
\begin{figure}[t]
  \centering
  \includegraphics[width=\columnwidth]{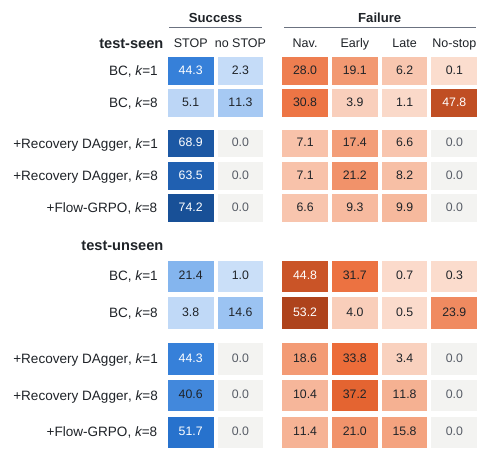}
  \caption{\textbf{Outcome attribution on the same recorded rollouts.} Each row splits a policy's episodes into two success and four failure types and sums to 100; values are expectations under the first-stop distribution (Sec.~\ref{sec:setup}), in \%. Color encodes magnitude within the success (blue) and failure (orange) groups. \emph{STOP} / \emph{no STOP}: success ending with an explicit in-region STOP / inside the region at the control budget. \emph{Nav.}: no in-region anchor even with STOP disabled; \emph{Early}/\emph{Late}: STOP outside the region before/after the first in-region anchor; \emph{No-stop}: reaches the region, never stops, and ends outside it. Rule in Appendix~\ref{sec:failure-rule}; two-decimal values in Table~\ref{tab:failure}.}
  \label{fig:failure}
\end{figure}

\begin{figure*}[t]
  \centering
  \includegraphics[width=\textwidth]{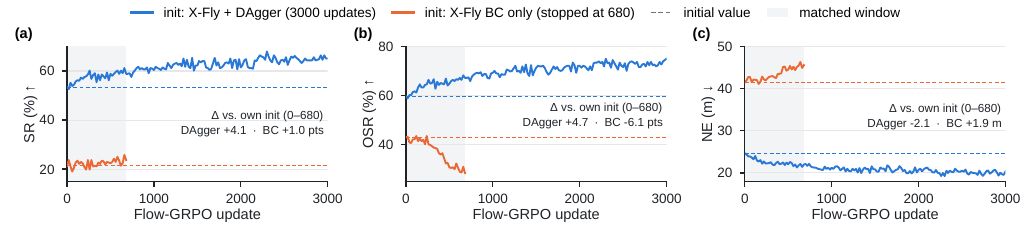}
  \caption{\textbf{Flow-GRPO from recovered vs.\ BC initialization.} Source-validation (a)~SR, (b)~OSR, and (c)~NE on a fixed 384-route BigCity panel excluded from Flow-GRPO updates, evaluated every 20 updates; dashed lines mark each initialization. The BC-initialized run was stopped at 680 updates as its OSR and NE kept degrading. $\Delta$: mean change of each run relative to its own initialization over the 35 matched evaluations (shaded window).}
  \label{fig:grpo-dynamics}
\end{figure*}

SR does not reveal whether an episode failed to reach the goal or failed to terminate there. We therefore assign each first-stop outcome of a recorded control rollout to one of two success types or to exactly one failure (Fig.~\ref{fig:failure}), weighted by its first-stop probability. A navigation failure means that the STOP-disabled control rollout contains no in-region anchor; termination failures (early, late, or no stop) are categorized on rollouts that do. This analysis separates control-rollout reachability from termination outcomes.

\paragraph{On test-seen, full-chunk BC mainly loses termination.}
On test-seen, BC fails to reach the region about as often at $k=8$ as at $k=1$ (30.77\% vs.\ 28.02\%), yet no-stop failures account for 47.78\% of its full-chunk outcomes, most of the SR drop from 46.57\% to 16.45\%. On test-unseen, full-chunk BC degrades on both sides: navigation failures rise from 44.80\% to 53.22\% and no-stop failures from 0.33\% to 23.88\%. Single-step BC instead mostly stops early (19.13\% seen, 31.74\% unseen). The STOP head, whose positive labels come only from expert terminal windows, does fire at single-step anchors, often too early, but rarely at the anchors produced by full-chunk execution.

\paragraph{Recovery changes both reaching and stopping failures.}
At $k=8$, Recovery DAgger reduces navigation failures from 30.77\% to 7.14\% on test-seen and from 53.22\% to 10.40\% on test-unseen, and no-stop failures fall to 0.00\% on both splits, consistent with STOP now being labeled at the learner's arrival anchors. Its unseen navigation failures are lower at $k=8$ than at $k=1$ (10.40\% vs.\ 18.56\%), consistent with labels being collected under the deployed execution. On test-seen, recovery narrows the $k=1$/$k=8$ SR difference from 30.12 to 5.42 points; on test-unseen the difference was already small (4.06 vs.\ 3.64 points), since there BC failed mostly at navigation under both horizons. Recovery raises unseen SR by similar amounts at both horizons (+21.78 at $k{=}1$, +22.20 at $k{=}8$), so on this split much of its gain is general correction at learner-visited states; the effect specific to full-chunk execution is concentrated on test-seen, where it removes the no-stop failures. The remaining full-chunk failures are dominated by early termination (21.21\% seen, 37.18\% unseen).

\paragraph{Flow-GRPO reduces premature stops.}
Relative to the $k=8$ recovery policy, Flow-GRPO keeps navigation failures at the recovered level (7.14$\rightarrow$6.59\% seen, 10.40$\rightarrow$11.39\% unseen) and roughly halves early termination (21.21$\rightarrow$9.34\% seen, 37.18$\rightarrow$21.04\% unseen), adding 10.72 and 11.10 SR points. The gain of this stage thus lies in termination timing rather than in control-rollout reachability. Its OSR gains (+11.20 seen, +15.36 unseen) largely follow from the same change, since fewer prefixes are truncated before the region. Late termination increases (8.19$\rightarrow$9.89\% seen, 11.80$\rightarrow$15.84\% unseen), consistent with Flow-GRPO shifting stops later. Termination dominates the residual failures, with late stops slightly more common than early stops on test-seen and early stops still more frequent on test-unseen. With STOP disabled, the final policy visits an in-region anchor in 93.41\% of test-seen and 88.61\% of test-unseen episodes, and on test-unseen dense reach exceeds anchor reach by only 0.75 points, so the remaining gap to SR lies mainly in stop timing at anchors the policy does visit.

\paragraph{Successes end with STOP after recovery.}
The benchmark also counts a rollout that ends inside the region without STOP as a success (column ``no STOP'' in Fig.~\ref{fig:failure}). For full-chunk BC, such successes make up 69.0\% of SR on test-seen and 79.4\% on test-unseen. After Recovery DAgger, at both horizons, and after Flow-GRPO, their share is 0.00\% on both splits, so every success of these policies ends with an explicit in-region STOP.

\paragraph{Effect of STOP opportunity count.}
Full-chunk execution also offers STOP only once every eight controls. To separate this effect, we replay the saved $k{=}1$ BC rollouts unchanged but permit STOP only every eight controls. SR becomes 32.10\% on test-seen and 32.56\% on test-unseen, compared with 46.57\% and 22.49\% for native $k{=}1$ BC and 16.45\% and 18.43\% for native $k{=}8$ BC. The effect of opportunity count differs between splits, and in both it leaves native full-chunk BC well below the replay, which keeps the $k{=}1$ visited states.

\subsection{Flow-GRPO training dynamics}
\label{sec:dynamics}
We track the Flow-GRPO run behind Table~\ref{tab:main}, started from the Recovery DAgger policy, together with the same procedure started from the BC policy. Both runs use a fixed source-validation panel of 384 BigCity routes excluded from Flow-GRPO updates, evaluated every 20 updates with fixed seeds and one sampled STOP rollout per route.

\paragraph{Improvement from recovered behavior.}
From the Recovery DAgger policy (Fig.~\ref{fig:grpo-dynamics}, blue), SR rises from 53.4\% to 65.1\% at update 3000, OSR from 59.6\% to 75.0\%, and NE falls from 24.7\,m to 20.3\,m. Improvement is fastest early, with SR first exceeding 60\% at update 260, and the three metrics improve together across 151 evaluations without a late reversal.

\paragraph{Initialization comparison.}
From the BC policy (orange), SR stays between 19.0\% and 25.8\% over 680 updates, while OSR declines from 43.0\% to 28.4\% and NE grows from 41.5\,m to 45.6\,m. Relative to its own initialization over the same 680 updates, the recovery-initialized run gains 4.1 SR and 4.7 OSR points and lowers NE by 2.1\,m, whereas the BC-initialized run gains 1.0 SR point, loses 6.1 OSR points, and raises NE by 1.9\,m. These matched-window results support recovery as an effective initialization for the tested post-training recipe.

\section{Limitations}
Termination timing remains the main residual bottleneck: the final policy visits an in-region anchor in most episodes but converts fewer of them into success. Each training configuration uses one training seed, so our results do not measure training-seed variance. OpenFly does not capture real aircraft dynamics, sensing noise, or operational safety constraints, and transfer to physical platforms requires separate validation.

\section{Conclusion}
We presented \method, a staged training framework for full-chunk aerial vision-language navigation. Recovery DAgger supplies learner-visited decision anchors with future-reconnecting chunks and arrival STOP labels, and Flow-GRPO jointly refines continuous chunked control and Bernoulli termination from same-task closed-loop outcomes. With eight controls per decision, \method reaches 74.18\% SR on OpenFly test-seen and 51.73\% on test-unseen at 4.7$\times$ less amortized model inference time per executed waypoint than single-step execution. Failure attribution identifies stop timing as the principal remaining challenge.

{\small\bibliographystyle{ieeenat_fullname}\bibliography{references}}

\begin{thebibliography}{35}
\providecommand{\natexlab}[1]{#1}
\providecommand{\url}[1]{\texttt{#1}}
\expandafter\ifx\csname urlstyle\endcsname\relax
  \providecommand{\doi}[1]{doi: #1}\else
  \providecommand{\doi}{doi: \begingroup \urlstyle{rm}\Url}\fi

\bibitem[Anderson et~al.(2018)Anderson, Wu, Teney, Bruce, Johnson,
  S{\"u}nderhauf, Reid, Gould, and van~den Hengel]{anderson2018vln}
Peter Anderson, Qi Wu, Damien Teney, Jake Bruce, Mark Johnson, Niko
  S{\"u}nderhauf, Ian Reid, Stephen Gould, and Anton van~den Hengel.
\newblock {Vision-and-Language Navigation: Interpreting Visually-Grounded
  Navigation Instructions in Real Environments}.
\newblock In \emph{Proceedings of the IEEE Conference on Computer Vision and
  Pattern Recognition (CVPR)}, pages 3674--3683, 2018.

\bibitem[Bansal et~al.(2019)Bansal, Krizhevsky, and
  Ogale]{bansal2019chauffeurnet}
Mayank Bansal, Alex Krizhevsky, and Abhijit Ogale.
\newblock {ChauffeurNet: Learning to Drive by Imitating the Best and
  Synthesizing the Worst}.
\newblock In \emph{Proceedings of Robotics: Science and Systems (RSS)}, 2019.

\bibitem[Black et~al.(2024)Black, Janner, Du, Kostrikov, and
  Levine]{black2024ddpo}
Kevin Black, Michael Janner, Yilun Du, Ilya Kostrikov, and Sergey Levine.
\newblock {Training Diffusion Models with Reinforcement Learning}.
\newblock In \emph{International Conference on Learning Representations
  (ICLR)}, pages 4965--4987, 2024.

\bibitem[Black et~al.(2025{\natexlab{a}})Black, Brown, Driess, Esmail, Equi,
  Finn, Fusai, Groom, Hausman, Ichter, Jakubczak, Jones, Ke, Levine, Li-Bell,
  Mothukuri, Nair, Pertsch, Shi, Tanner, Vuong, Walling, Wang, and
  Zhilinsky]{black2025pi0}
Kevin Black, Noah Brown, Danny Driess, Adnan Esmail, Michael Equi, Chelsea
  Finn, Niccolo Fusai, Lachy Groom, Karol Hausman, Brian Ichter, Szymon
  Jakubczak, Tim Jones, Liyiming Ke, Sergey Levine, Adrian Li-Bell, Mohith
  Mothukuri, Suraj Nair, Karl Pertsch, Lucy~Xiaoyang Shi, James Tanner, Quan
  Vuong, Anna Walling, Haohuan Wang, and Ury Zhilinsky.
\newblock {$\pi_0$: A Vision-Language-Action Flow Model for General Robot
  Control}.
\newblock In \emph{Proceedings of Robotics: Science and Systems (RSS)},
  2025{\natexlab{a}}.

\bibitem[Black et~al.(2025{\natexlab{b}})Black, Galliker, and
  Levine]{black2025rtc}
Kevin Black, Manuel Galliker, and Sergey Levine.
\newblock {Real-Time Execution of Action Chunking Flow Policies}.
\newblock In \emph{Advances in Neural Information Processing Systems}, pages
  33383--33407, 2025{\natexlab{b}}.

\bibitem[Chen et~al.(2021)Chen, Guhur, Schmid, and Laptev]{chen2021hamt}
Shizhe Chen, Pierre-Louis Guhur, Cordelia Schmid, and Ivan Laptev.
\newblock {History Aware Multimodal Transformer for Vision-and-Language
  Navigation}.
\newblock In \emph{Advances in Neural Information Processing Systems}, pages
  5834--5847, 2021.

\bibitem[Chen et~al.(2022)Chen, Guhur, Tapaswi, Schmid, and
  Laptev]{chen2022duet}
Shizhe Chen, Pierre-Louis Guhur, Makarand Tapaswi, Cordelia Schmid, and Ivan
  Laptev.
\newblock {Think Global, Act Local: Dual-Scale Graph Transformer for
  Vision-and-Language Navigation}.
\newblock In \emph{Proceedings of the IEEE/CVF Conference on Computer Vision
  and Pattern Recognition (CVPR)}, pages 16537--16547, 2022.

\bibitem[Chi et~al.(2025)Chi, Xu, Feng, Cousineau, Du, Burchfiel, Tedrake, and
  Song]{chi2024diffusionpolicy}
Cheng Chi, Zhenjia Xu, Siyuan Feng, Eric Cousineau, Yilun Du, Benjamin
  Burchfiel, Russ Tedrake, and Shuran Song.
\newblock {Diffusion Policy: Visuomotor Policy Learning via Action Diffusion}.
\newblock \emph{The International Journal of Robotics Research}, 44\penalty0
  (10--11):\penalty0 1684--1704, 2025.

\bibitem[Deng and Xu(2026)]{deng2026dreamfly}
Yan Deng and Fei Xu.
\newblock {DreamFly: Causal Memory and Receding-Horizon Diffusion Planning for
  Aerial Vision-Language Navigation}.
\newblock \emph{arXiv preprint arXiv:2608.12308}, 2026.

\bibitem[Fried et~al.(2018)Fried, Hu, Cirik, Rohrbach, Andreas, Morency,
  Berg-Kirkpatrick, Saenko, Klein, and Darrell]{fried2018speaker}
Daniel Fried, Ronghang Hu, Volkan Cirik, Anna Rohrbach, Jacob Andreas,
  Louis-Philippe Morency, Taylor Berg-Kirkpatrick, Kate Saenko, Dan Klein, and
  Trevor Darrell.
\newblock {Speaker-Follower Models for Vision-and-Language Navigation}.
\newblock In \emph{Advances in Neural Information Processing Systems}, 2018.

\bibitem[Gao et~al.(2026)Gao, Li, You, Liu, Li, Chen, Chen, Tang, Wang, Yang,
  Tang, Tang, Liang, Zhu, Xiong, Su, Ye, Li, Ding, Wang, Li, Wang, and
  Zhao]{gao2025openfly}
Yunpeng Gao, Chenhui Li, Zhongrui You, Junli Liu, Zhen Li, Pengan Chen, Qizhi
  Chen, Zhonghan Tang, Liansheng Wang, Penghui Yang, Yiwen Tang, Yuhang Tang,
  Shuai Liang, Songyi Zhu, Ziqin Xiong, Yifei Su, Xinyi Ye, Jianan Li, Yan
  Ding, Dong Wang, Xuelong Li, Zhigang Wang, and Bin Zhao.
\newblock {OpenFly: A Comprehensive Platform for Aerial Vision-Language
  Navigation}.
\newblock In \emph{International Conference on Learning Representations
  (ICLR)}, pages 101858--101871, 2026.

\bibitem[Hong et~al.(2021)Hong, Wu, Qi, Rodriguez-Opazo, and
  Gould]{hong2021vlnbert}
Yicong Hong, Qi Wu, Yuankai Qi, Cristian Rodriguez-Opazo, and Stephen Gould.
\newblock {VLN BERT: A Recurrent Vision-and-Language BERT for Navigation}.
\newblock In \emph{Proceedings of the IEEE/CVF Conference on Computer Vision
  and Pattern Recognition (CVPR)}, pages 1643--1653, 2021.

\bibitem[Kelly et~al.(2019)Kelly, Sidrane, Driggs-Campbell, and
  Kochenderfer]{kelly2019hgdagger}
Michael Kelly, Chelsea Sidrane, Katherine Driggs-Campbell, and Mykel~J.
  Kochenderfer.
\newblock {HG-DAgger: Interactive Imitation Learning with Human Experts}.
\newblock In \emph{2019 International Conference on Robotics and Automation
  (ICRA)}, pages 8077--8083, 2019.

\bibitem[Krantz et~al.(2020)Krantz, Wijmans, Majumdar, Batra, and
  Lee]{krantz2020beyond}
Jacob Krantz, Erik Wijmans, Arjun Majumdar, Dhruv Batra, and Stefan Lee.
\newblock {Beyond the Nav-Graph: Vision-and-Language Navigation in Continuous
  Environments}.
\newblock In \emph{Computer Vision -- ECCV 2020}, pages 104--120. Springer,
  2020.

\bibitem[Laskey et~al.(2017)Laskey, Lee, Fox, Dragan, and
  Goldberg]{laskey2017dart}
Michael Laskey, Jonathan Lee, Roy Fox, Anca Dragan, and Ken Goldberg.
\newblock {DART: Noise Injection for Robust Imitation Learning}.
\newblock In \emph{Proceedings of the 1st Annual Conference on Robot Learning
  (CoRL)}, pages 143--156. PMLR, 2017.

\bibitem[Li et~al.(2026)Li, Gong, Thompson, Javanmardi, Zhang, Zeng, Liu, Wang,
  Guo, and Tsukada]{li2026think}
Yun Li, Jiachen Gong, Simon Thompson, Ehsan Javanmardi, Qunli Zhang, Zifan
  Zeng, Shiming Liu, Peng Wang, Zixuan Guo, and Manabu Tsukada.
\newblock {Think at 5 Hz, Act at 20 Hz: Asynchronous Fast-Slow
  Vision-Language-Action Inference for Closed-Loop Driving}.
\newblock \emph{arXiv preprint arXiv:2607.15621}, 2026.

\bibitem[Lipman et~al.(2023)Lipman, Chen, Ben-Hamu, Nickel, and
  Le]{lipman2023flow}
Yaron Lipman, Ricky T.~Q. Chen, Heli Ben-Hamu, Maximilian Nickel, and Matt Le.
\newblock {Flow Matching for Generative Modeling}.
\newblock In \emph{International Conference on Learning Representations
  (ICLR)}, 2023.

\bibitem[Liu et~al.(2025{\natexlab{a}})Liu, Liu, Liang, Li, Liu, Wang, Wan,
  Zhang, and Ouyang]{liu2025flowgrpo}
Jie Liu, Gongye Liu, Jiajun Liang, Yangguang Li, Jiaheng Liu, Xintao Wang,
  Pengfei Wan, Di Zhang, and Wanli Ouyang.
\newblock {Flow-GRPO: Training Flow Matching Models via Online RL}.
\newblock In \emph{Advances in Neural Information Processing Systems}, pages
  40783--40818, 2025{\natexlab{a}}.

\bibitem[Liu et~al.(2023)Liu, Zhang, Qi, Wang, Zhang, and Wu]{liu2023aerialvln}
Shubo Liu, Hongsheng Zhang, Yuankai Qi, Peng Wang, Yanning Zhang, and Qi Wu.
\newblock {AerialVLN: Vision-and-Language Navigation for UAVs}.
\newblock In \emph{Proceedings of the IEEE/CVF International Conference on
  Computer Vision (ICCV)}, pages 15384--15394, 2023.

\bibitem[Liu et~al.(2026)Liu, Huang, Xia, Liu, Cui, and
  Yang]{liu2026imagineuav}
Xuchen Liu, Jiawei Huang, Shihao Xia, Bingxi Liu, Jinqiang Cui, and Jiankun
  Yang.
\newblock {ImagineUAV: Aerial Vision-Language Navigation via World-Action
  Modeling and Kinodynamic Planning}.
\newblock \emph{arXiv preprint arXiv:2606.01205}, 2026.

\bibitem[Liu et~al.(2025{\natexlab{b}})Liu, Hamid, Xie, Lee, Du, and
  Finn]{liu2025bid}
Yuejiang Liu, Jubayer~Ibn Hamid, Annie Xie, Yoonho Lee, Maximilian Du, and
  Chelsea Finn.
\newblock {Bidirectional Decoding: Improving Action Chunking via Guided
  Test-Time Sampling}.
\newblock In \emph{International Conference on Learning Representations
  (ICLR)}, pages 4594--4627, 2025{\natexlab{b}}.

\bibitem[Ma et~al.(2019)Ma, Lu, Wu, AlRegib, Kira, Socher, and
  Xiong]{ma2019selfmonitoring}
Chih-Yao Ma, Jiasen Lu, Zuxuan Wu, Ghassan AlRegib, Zsolt Kira, Richard Socher,
  and Caiming Xiong.
\newblock {Self-Monitoring Navigation Agent via Auxiliary Progress Estimation}.
\newblock In \emph{International Conference on Learning Representations
  (ICLR)}, 2019.

\bibitem[Ren et~al.(2025)Ren, Lidard, Ankile, Simeonov, Agrawal, Majumdar,
  Burchfiel, Dai, and Simchowitz]{ren2025dppo}
Allen~Z. Ren, Justin Lidard, Lars~L. Ankile, Anthony Simeonov, Pulkit Agrawal,
  Anirudha Majumdar, Benjamin Burchfiel, Hongkai Dai, and Max Simchowitz.
\newblock {Diffusion Policy Policy Optimization}.
\newblock In \emph{International Conference on Learning Representations
  (ICLR)}, pages 77288--77329, 2025.

\bibitem[Ross et~al.(2011)Ross, Gordon, and Bagnell]{ross2011dagger}
St{\'e}phane Ross, Geoffrey Gordon, and Drew Bagnell.
\newblock {A Reduction of Imitation Learning and Structured Prediction to
  No-Regret Online Learning}.
\newblock In \emph{Proceedings of the Fourteenth International Conference on
  Artificial Intelligence and Statistics (AISTATS)}, pages 627--635. PMLR,
  2011.

\bibitem[Tan et~al.(2019)Tan, Yu, and Bansal]{tan2019envdrop}
Hao Tan, Licheng Yu, and Mohit Bansal.
\newblock {Learning to Navigate Unseen Environments: Back Translation with
  Environmental Dropout}.
\newblock In \emph{Proceedings of the 2019 Conference of the North American
  Chapter of the Association for Computational Linguistics: Human Language
  Technologies, Volume 1 (Long and Short Papers)}, pages 2610--2621.
  Association for Computational Linguistics, 2019.

\bibitem[Wang et~al.(2026)Wang, Kang, Sun, Li, Yu, and
  Tian]{wang2026recoverfly}
Boxiong Wang, Hui Kang, Geng Sun, Jiahui Li, Chao Yu, and Daxin Tian.
\newblock {RefineFly: Failure-Aware Post-Training for Aerial Vision-Language
  Navigation}.
\newblock \emph{arXiv preprint arXiv:2608.09467}, 2026.

\bibitem[Wang et~al.(2019)Wang, Huang, Celikyilmaz, Gao, Shen, Wang, Wang, and
  Zhang]{wang2019rcm}
Xin Wang, Qiuyuan Huang, Asli Celikyilmaz, Jianfeng Gao, Dinghan Shen,
  Yuan-Fang Wang, William~Yang Wang, and Lei Zhang.
\newblock {Reinforced Cross-Modal Matching and Self-Supervised Imitation
  Learning for Vision-Language Navigation}.
\newblock In \emph{Proceedings of the IEEE/CVF Conference on Computer Vision
  and Pattern Recognition (CVPR)}, pages 6629--6638, 2019.

\bibitem[Wang et~al.(2025)Wang, Yang, Wang, Kwan, Chen, Wu, Li, Liao, and
  Liu]{wang2025traveluav}
Xiangyu Wang, Donglin Yang, Ziqin Wang, Hohin Kwan, Jinyu Chen, Wenjun Wu,
  Hongsheng Li, Yue Liao, and Si Liu.
\newblock {Towards Realistic UAV Vision-Language Navigation: Platform,
  Benchmark, and Methodology}.
\newblock In \emph{International Conference on Learning Representations
  (ICLR)}, pages 7292--7310, 2025.

\bibitem[Xiang et~al.(2020)Xiang, Wang, and Wang]{xiang2020stop}
Jiannan Xiang, Xin Wang, and William~Yang Wang.
\newblock {Learning to Stop: A Simple yet Effective Approach to Urban
  Vision-Language Navigation}.
\newblock In \emph{Findings of the Association for Computational Linguistics:
  EMNLP 2020}, pages 699--707. Association for Computational Linguistics, 2020.

\bibitem[Xu et~al.(2025)Xu, Wang, Xia, Zhu, Huang, and Xu]{xu2025vlacache}
Siyu Xu, Yunke Wang, Chenghao Xia, Dihao Zhu, Tao Huang, and Chang Xu.
\newblock {VLA-Cache: Efficient Vision-Language-Action Manipulation via
  Adaptive Token Caching}.
\newblock In \emph{Advances in Neural Information Processing Systems}, pages
  164448--164473, 2025.

\bibitem[Zhang et~al.(2025{\natexlab{a}})Zhang, Yu, Su, and
  Wang]{zhang2025reinflow}
Tonghe Zhang, Chao Yu, Sichang Su, and Yu Wang.
\newblock {ReinFlow: Fine-tuning Flow Matching Policy with Online Reinforcement
  Learning}.
\newblock In \emph{Advances in Neural Information Processing Systems}, pages
  106282--106319, 2025{\natexlab{a}}.

\bibitem[Zhang et~al.(2025{\natexlab{b}})Zhang, Gao, Yu, Peng, Zhao, Zhang,
  Cui, Chen, and Li]{zhang2025citynavagent}
Weichen Zhang, Chen Gao, Shiquan Yu, Ruiying Peng, Baining Zhao, Qian Zhang,
  Jinqiang Cui, Xinlei Chen, and Yong Li.
\newblock {CityNavAgent: Aerial Vision-and-Language Navigation with
  Hierarchical Semantic Planning and Global Memory}.
\newblock In \emph{Proceedings of the 63rd Annual Meeting of the Association
  for Computational Linguistics (Volume 1: Long Papers)}, pages 31292--31309.
  Association for Computational Linguistics, 2025{\natexlab{b}}.

\bibitem[Zhao et~al.(2026)Zhao, Xu, Feng, Zhang, Wang, Wang, Ji, Wang, Fang,
  Zheng, Zhang, Shang, Wu, Gao, Chen, and Li]{zhao2026worldvln}
Baining Zhao, Jiacheng Xu, Weicheng Feng, Xin Zhang, Zhaolu Wang, Haoyang Wang,
  Shilong Ji, Ziyou Wang, Jianjie Fang, Zhiheng Zheng, Weichen Zhang, Yu Shang,
  Wei Wu, Chen Gao, Xinlei Chen, and Yong Li.
\newblock {WorldVLN: Autoregressive World Action Model for Aerial
  Vision-Language Navigation}.
\newblock \emph{arXiv preprint arXiv:2605.15964}, 2026.

\bibitem[Zhao et~al.(2023)Zhao, Kumar, Levine, and Finn]{zhao2023act}
Tony~Z. Zhao, Vikash Kumar, Sergey Levine, and Chelsea Finn.
\newblock {Learning Fine-Grained Bimanual Manipulation with Low-Cost Hardware}.
\newblock In \emph{Proceedings of Robotics: Science and Systems (RSS)}, 2023.

\bibitem[Zhu et~al.(2026)Zhu, Meng, Yu, Zhang, Ma, Zhou, and
  Tian]{zhu2026fsdvln}
Xueke Zhu, Qingyan Meng, Liutao Yu, Wei Zhang, Zhengyu Ma, Huihui Zhou, and
  Yonghong Tian.
\newblock {FSD-VLN: Fast-Slow Dual-System Modeling for Aerial Long-Horizon
  Vision-Language Navigation}.
\newblock \emph{arXiv preprint arXiv:2607.08359}, 2026.

\end{thebibliography}
\clearpage
\appendix
\section{Implementation Details}
\label{sec:impl}
\subsection{Training and execution pseudocode}
\label{sec:pseudocode}
\paragraph{Recovery-label construction.}
\begin{enumerate}[leftmargin=1.35em,itemsep=1pt,topsep=2pt]
  \item Roll out the learner with the deployed full-chunk executor; after each CONTINUE, advance the monotonic expert cursor by the number of controls actually executed.
  \item At anchor $s_t$, compute the learner-to-goal distance. If this is the first anchor within 20\,m, emit \textsc{Stop} and an all-zero hover target, then end collection for that route.
  \item Otherwise emit \textsc{Continue}. Clamp an overrun cursor to the final expert pose and construct the $H$-step target of Eq.~\ref{eq:recovery}; an outside-region cursor overrun therefore remains a CONTINUE example pointing toward the goal.
  \item Reduce the initial yaw offset to the shortest signed angle before interpolation and wrap every relative yaw target. Mix the retained recovery windows 1:1 with BC windows and train the DAgger stage.
\end{enumerate}

\paragraph{Flow-GRPO update.}
\begin{enumerate}[leftmargin=1.35em,itemsep=1pt,topsep=2pt]
  \item For each task, sample eight closed-loop attempts with shared initial flow noise: six free attempts and two STOP-override attempts, $n\sim\mathrm{Uniform}\{1,2,3\}$.
  \item At each legal anchor, sample a flow chunk and Bernoulli STOP. A STOP-override attempt overrides its first $n$ sampled STOP events to CONTINUE and then runs freely.
  \item Compute complete trajectory outcomes and depth-aligned group advantages. Override outcomes enter the group statistics, but override decisions are excluded from both actor surrogates.
  \item From STOP-override trajectories, form same-depth outcome-verified pairs: an overridden outside-region STOP is negative only if continuation later reaches a legal in-region anchor; a legal in-region decision is positive. Apply the rank+BCE loss and update the joint trainable policy.
\end{enumerate}

\begin{table}[h]
\centering
\caption{Data flow of the three training stages. Every stage trains the same modules; the VLM and visual tower stay frozen.}
\label{tab:stages}
\footnotesize
\setlength{\tabcolsep}{2.5pt}
\begin{tabular}{@{}p{0.16\columnwidth}p{0.25\columnwidth}p{0.28\columnwidth}p{0.23\columnwidth}@{}}
\toprule
Stage & Init.\ / data & Labels or signal & Length / selection\\
\midrule
BC & scratch; 849,106 expert windows (UE, 3DGS, AirSim) & expert chunks; STOP at terminal windows & 10 epochs\\
Recovery DAgger & BC, then previous-round checkpoint (3 rounds); learner rollouts at $k{=}8$, STOP disabled, mixed 1:1 with BC & recovery chunks (Eq.~\ref{eq:recovery}); STOP at first in-region anchor & 3 epochs per round; lowest val.\ control-speed error\\
Flow-GRPO & Recovery DAgger at $k{=}8$; BigCity training tasks & group-relative advantages (Eq.~\ref{eq:advantage}); outcome-verified STOP pairs & 3000 updates; val.\ panel score (App.~\ref{sec:impl})\\
\bottomrule
\end{tabular}
\end{table}

\subsection{Policy architecture and design choices}
\paragraph{Slow path.}
A frozen Qwen3.5-4B receives up to four causal keyframes, oldest first, followed by the instruction. Keyframes refresh on a fixed timer every 16 executed control steps rather than being triggered by predicted actions; their contents still reflect the executed trajectory. The VLM decoder re-runs only when the cache changes. We read the 2560-wide hidden states of the fourth-last decoder layer. Two learned query banks cross-attend to them: an observation bank restricted to image-token positions and a navigation bank restricted to the instruction span. Each bank holds 48 tokens of width 768 and applies one 12-head cross-attention.

\paragraph{Fast path.}
Only the visual tower runs on the current $448\times448$ frame. Its last-block patch tokens ($784\times1024$) are compressed by a third query bank into 48 current-frame tokens. Running the decoder at every decision would dominate inference cost, while refreshing only at keyframes would leave the policy without the current view between refreshes; the split keeps the current frame at visual-tower cost.

\paragraph{State and history.}
The 10-D ego state contains the last-step displacement in the body frame, sine and cosine of the last-step yaw change, speed, yaw rate, and the displacement from the start pose in the start frame. All entries are relative; no goal or global coordinate is provided. An anchor token encodes the pose change since the slow cache was formed and the normalized cache age, and a learned age embedding added to the observation tokens lets the policy discount or re-interpret stale keyframes. History tokens are built from the executed pose history; predicted but unexecuted controls are excluded, while executed actions affect both subsequent observations and history. Empty history slots are masked in attention, and BC drops history slots with probability 0.15.

\paragraph{Fusion.}
The main sequence concatenates five learned task tokens (action, waypoint, STOP, direction, chunk), the observation and current-frame tokens, the state and anchor tokens, and the history tokens, each with a type embedding. Each of the three fusion blocks applies pre-LN self-attention over this sequence, ungated cross-attention to the navigation tokens, and a feed-forward layer, with width 768, 12 heads, and dropout 0.1; a final LayerNorm follows.

\paragraph{Controller and STOP head.}
The DiT conditions its AdaLN-Zero modulation only on flow time and reads the fused tokens through ungated cross-attention; its output projection is zero-initialized. The controller is small (2 layers, width 256, 8 heads, 3.9M parameters) and is integrated with six Euler steps. The STOP head is a two-layer MLP on the fused STOP token. It is computed before and independently of the controller, so termination does not condition on the sampled chunk and STOP and control keep separate likelihoods. The trainable modules total about 45M parameters. The VLM and visual tower remain frozen across all three training stages.

\subsection{Behavior cloning}
BC uses AdamW (learning rate $10^{-4}$, weight decay 0.01) with 500 warmup steps and cosine decay, an effective batch of 128 on 8 GPUs, 10 epochs, bf16 autocast, and gradient clipping at 1. Besides the rectified-flow loss, the objective includes STOP binary cross-entropy (weight 0.1), an endpoint loss on the final position of the one-step chunk estimate (0.1), control smoothness and speed consistency (0.1 each), yaw continuity (0.05), and an auxiliary label-smoothed discrete-action cross-entropy on the fused action token (0.2), used only during BC. STOP positives are sampled annotation steps whose dense expert action is STOP. Expert chunks that extend past the trajectory end hold the final pose with zero motion. The three domains are sampled in their natural proportions.

\subsection{Recovery DAgger}
Collection, RL, and evaluation share the full-chunk executor; the route cursor advances by the number of controls actually executed. Recovery collection runs with STOP disabled and is truncated at the first in-region anchor; free RL rollouts sample the STOP policy, and evaluation uses its first-stop distribution (Sec.~\ref{sec:setup}); both follow the benchmark termination rules. Label construction then applies a geometric priority rule. The prefix is cut at the first learner anchor within 20\,m, which receives STOP and an all-zero hover target. If the learner never enters, all retained anchors remain CONTINUE even after the cursor passes the last expert pose; the clamped reference stays at the goal and the recovery target continues toward it. The collected set contains 8,404 recovery STOP positives. Only successfully rendered, finite anchors are retained. Before interpolation, the yaw offset is reduced modulo $2\pi$ to the shortest signed angle; each relative yaw control is wrapped again. There are three collection-and-training rounds: the first starts from BC, and each later round warm-starts from the previous best checkpoint with a fresh optimizer. Each round trains for 3 epochs at learning rate $3\times10^{-5}$ with 50 warmup steps, cosine decay, and batch 128; the auxiliary action loss is switched off. The 1:1 mixture pairs the recovery labels with a seeded random subsample of BC windows of equal size, and the checkpoint with the lowest validation control-speed error is kept.

\subsection{Flow-GRPO}
\paragraph{Recipe.}
For late-depth advantage normalization, sparse depths with fewer than four active attempts pool the next adjacent depth; a depth without a valid four-sample comparison receives zero advantage. Route difficulty is computed from static geometry as a weighted sum of percentile-ranked route length (0.50), accumulated turning (0.35), and vertical travel (0.15). Each training panel mixes 50\% natural-distribution coverage, 20\% curriculum samples, and 30\% current-policy-prioritized samples. The curriculum first draws from the easiest 45\% of routes and expands to 70\% after 108 updates, and prioritized sampling rises to 50\% over all routes after update 324; priority scores favor tasks with informative reward spread and mixed success while penalizing repeated exposure. We use six flow integration steps, linear schedule endpoints $\sigma_{\min}=0.05$ and $\sigma_{\max}=0.20$, three stochastic tail steps, four temporal exploration modes, complement-noise ratio $\rho=0.1$, and one optimizer step per rollout batch. The gradient norm is clipped at 1, and both reference-KL weights are 0.01. Goal progress is clipped to 0.4 per decision and 2.5 per episode; invalid, direction, and smoothness penalties have episode caps of 2, 2, and 4. The largest positive shaping an unsuccessful episode can collect is the first-entry bonus plus the progress cap, $2+2.5=4.5$, so we set the failure shift to $\kappa=4.51$. An early STOP outside the region receives no penalty beyond this shift. Training runs for 3000 updates.

\paragraph{Sampling.}
Each update draws 48 BigCity training tasks with goal distance at least 20\,m and at least 10 expert steps, run as six groups on each of 8 GPUs. Within a group, attempts share the initial flow latent at each decision depth, seeded by update, task, and depth; SDE-tail noise and STOP draws use per-attempt seeds. STOP is legal from the second decision on and is sampled from $\sigma(\ell/1.5)$ during training rollouts to widen exploration of stop timing. This temperature applies only to Flow-GRPO training; BC and Recovery DAgger train the STOP head at unit temperature, and all reported evaluations use $\sigma(\ell)$.

\paragraph{Counterfactual attempts and credit.}
Two attempts per group are STOP-override attempts with $n$ drawn uniformly from $\{1,2,3\}$: each of their first $n$ sampled STOPs is overridden to CONTINUE, after which the attempt runs freely. They are complete rollouts from the shared start. Their returns enter the depth-wise statistics of Eq.~\ref{eq:advantage}, but none of their decisions enters either term of Eq.~\ref{eq:actor}, so the policy-gradient terms come from the six free attempts. Override returns do not depend on a free attempt's own SDE-tail and STOP noise, so including them in the depth-wise mean does not bias that attempt's gradient, and including its own return only rescales its centered advantage by $1-1/|\mathcal A_t|$. The update is nevertheless not an unbiased gradient estimate: the standard deviation, a population standard deviation over all attempts at a depth, depends on the attempt's own return, and pooling by depth conditions on which attempts survive to that depth. Outcome terms (task, first entry, leaving, failure shift) are added undiscounted to every decision's return-to-go, and shaping terms are discounted by 0.99 per executed waypoint.

\paragraph{Reward terms.}
With $\phi(d)=\max(d-20,0)$, goal progress adds $0.03(\phi_{j-1}-\phi_j)$ per executed waypoint before the caps listed above. A direction penalty applies to horizontal steps of at least 0.25\,m whose motion deviates more than $60^\circ$ from the current yaw: $0.08$ per meter times a severity that rises linearly from 0 at $60^\circ$ to 1 at $90^\circ$ and continues linearly beyond. Smoothness costs $0.06$ times the relative velocity change and $0.02$ times the normalized yaw-rate change per waypoint, capped at 0.35 per decision. A decision with non-finite waypoints costs 0.8, and a non-finite pose or render error removes the group from the update. First entry is detected at 20\,m and leaving at 22\,m. An episode that ends inside the region without STOP counts as successful and receives no terminal reward or failure shift.

\paragraph{Objective.}
The control KL is the exact Gaussian divergence between the current and reference restoring means under the shared covariance of Eq.~\ref{eq:sde}, restricted to executed waypoints and summed over the stochastic steps; the STOP KL is the Bernoulli divergence at legal free decisions. Likelihood and KL terms are normalized by the number of environment decisions in the update. A control term is active only when the decision executes a chunk prefix, and no decision after STOP receives credit. AdamW without weight decay uses learning rate $1.5\times10^{-6}$ for the flow controller and $5\times10^{-7}$ for all other trainable modules, including fusion and the STOP head, with one optimizer step per update.

\paragraph{Ranking term.}
Pairs are formed among STOP-override decisions at the same depth within a GPU's batch. A positive is a legal decision within 20\,m of the goal; a negative is an overridden STOP more than 22\,m from the goal whose attempt later reaches a legal in-region decision. At most two pairs are kept per depth. With temperature-scaled logits $\tilde\ell=\ell/1.5$,
\begin{equation}
\begin{split}
\mathcal L_{\mathrm{rank}}={}&\operatorname{softplus}\!\bigl(1-(\tilde\ell^{+}-\tilde\ell^{-})\bigr)\\
&+\tfrac12\bigl[\operatorname{softplus}(-\tilde\ell^{+})+\operatorname{softplus}(\tilde\ell^{-})\bigr],
\end{split}
\end{equation}
averaged over pairs and added to the actor loss with weight 0.5.

\paragraph{Validation and selection.}
The validation panel holds 384 fixed BigCity training-split routes excluded from the task manifest used for Flow-GRPO updates and is evaluated every 20 updates with a fixed seed and one Bernoulli STOP rollout per route. A checkpoint is scored by the mean of its last three valid evaluations. Candidates whose OSR does not fall more than 2 points below the initial OSR are ranked by SR, then by a smaller OSR--SR gap, then by lower final distance. The test splits are not used for selection.

\begin{table*}[t]
\centering
\caption{Outcome attribution (\% of episodes) underlying Fig.~\ref{fig:failure}. Labels follow the rule above. The two success columns sum to SR (Table~\ref{tab:main}), and all six columns of a split sum to 100 up to rounding.}
\label{tab:failure}
\resizebox{\textwidth}{!}{\begin{tabular}{lrrrrrrrrrrrr}
\toprule
& \multicolumn{6}{c}{test-seen} & \multicolumn{6}{c}{test-unseen}\\
\cmidrule(lr){2-7}\cmidrule(lr){8-13}
& \multicolumn{2}{c}{Success} & & & & & \multicolumn{2}{c}{Success} & & & & \\
\cmidrule(lr){2-3}\cmidrule(lr){8-9}
Method & STOP & no STOP & Nav. & Early & Late & No-stop & STOP & no STOP & Nav. & Early & Late & No-stop\\
\midrule
BC, $k{=}1$ & 44.27 & 2.30 & 28.02 & 19.13 & 6.20 & 0.07 & 21.45 & 1.04 & 44.80 & 31.74 & 0.65 & 0.33\\
BC, $k{=}8$ & 5.10 & 11.35 & 30.77 & 3.85 & 1.15 & 47.78 & 3.80 & 14.63 & 53.22 & 3.98 & 0.48 & 23.88\\
+ Recovery DAgger, $k{=}1$ & 68.88 & 0.00 & 7.14 & 17.41 & 6.57 & 0.00 & 44.27 & 0.00 & 18.56 & 33.80 & 3.37 & 0.00\\
+ Recovery DAgger, $k{=}8$ & 63.46 & 0.00 & 7.14 & 21.21 & 8.19 & 0.00 & 40.63 & 0.00 & 10.40 & 37.18 & 11.80 & 0.00\\
+ Flow-GRPO, $k{=}8$ & 74.18 & 0.00 & 6.59 & 9.34 & 9.89 & 0.00 & 51.73 & 0.00 & 11.39 & 21.04 & 15.84 & 0.00\\
\bottomrule
\end{tabular}}
\end{table*}

\subsection{Evaluation}
Each decision integrates the flow ODE with six deterministic Euler steps from a fresh Gaussian latent, seeded per route. Each route is recorded once with STOP disabled at the given $k$, and the first-stop metrics of Sec.~\ref{sec:setup} are computed from the stored STOP logits at unit temperature, $p_t=\sigma(\ell_t)$; the same recording yields anchor and dense reach and the failure attribution of Appendix~\ref{sec:failure-rule}. All \method rows use the same cap of 128 executed controls, independent of $k$, together with the same executor, success radius, and termination rule. Each first-stop outcome is scored on its own executed prefix, so a prefix that stops before a later abnormal event keeps its metrics. A trajectory that ends abnormally (collision, render failure, or non-finite state) before any replayable decision contributes zero to SR, OSR, and SPL and remains in the denominator. SPL uses the executed path length up to the stop, and metrics are averaged over all routes. Prior-work rows are their published benchmark results with native stopping policies.

\subsection{Training dynamics}
The BC-initialized run in Sec.~\ref{sec:dynamics} uses this recipe unchanged, with the BC policy as initialization and KL reference. Validation points interrupted by renderer failures were evaluated from the saved checkpoints with the same panel and seeds.

\section{Transition Density}
\label{sec:density}
In Eq.~\ref{eq:sde}, $\Delta t=1/6$, $\nu_s=\sigma_s\sqrt{\Delta t}$, and $\sigma_s=0.20-0.15s/6$. The restoring mean uses the score implied by the velocity field,
\begin{equation}
\mu_\theta^s=x^s+\Delta t\,v_\theta(x^s,t_s,c)-\tfrac{1}{2(1-t_s)}\,C_s\bigl(x^s-t_sv_\theta(x^s,t_s,c)\bigr),
\label{eq:restoring}
\end{equation}
so that drift and noise share the same anisotropic covariance, and the small complement noise gives old and updated policies common support. For a stochastic step of Eq.~\ref{eq:sde}, let $e_s=x^{s+1}-\mu_\theta^s$, $a_s=\Phi^\top e_s$, and $b_s=\Psi^\top e_s$. Since $C_s$ is diagonal in the $(\Phi,\Psi)$ basis,
\begin{equation}
\begin{split}
\log \pi_\theta^s=-\frac12\sum_{d=1}^{D}\Bigl\{&\sum_{m=1}^{M}\bigl[a_{smd}^2/\nu_s^2+\log(2\pi\nu_s^2)\bigr]\\
&+\sum_{n=1}^{H-M}\bigl[b_{snd}^2/(\rho^2\nu_s^2)+\log(2\pi\rho^2\nu_s^2)\bigr]\Bigr\}.
\end{split}
\label{eq:mode-logp}
\end{equation}
This scores all low-frequency and complement coordinates jointly, \ie the complete $H\times D$ transition, rather than treating waypoints as independent actions. The three deterministic steps carry no density and enter no likelihood term. As in Flow-GRPO, their output $x^3$ is stored as part of the state. With $g_\theta^s=-(x^s-t_sv_\theta)/(1-t_s)$, the score of the interpolant $X_\tau=(1-\tau)\epsilon+\tau U$ under the flow's own prediction, Eq.~\ref{eq:restoring} reads $\mu_\theta^s=x^s+\Delta t\,v_\theta+\tfrac12C_sg_\theta^s$.

\section{Outcome Attribution}
\label{sec:failure-rule}
For a control rollout with STOP disabled, let $t^\star$ be its first decision anchor inside the 20\,m region ($t^\star=\infty$ if none). A first-stop outcome at decision $t$ (or at the control budget if no STOP occurs) that ends inside the region is a success, with an explicit STOP if it ends at a STOP decision and without STOP if it ends at the control budget. Otherwise it is labeled \emph{Nav.}\ if $t^\star=\infty$, \emph{No-stop} if no STOP occurs, \emph{Early} if $t<t^\star$, and \emph{Late} if $t>t^\star$, checked in this order. Each label is weighted by the first-stop probability of its outcome.

\end{document}